\documentclass[10pt, a4paper]{article}

\usepackage{lrec-coling2024} 
\usepackage{times}
\usepackage{latexsym}
\usepackage{amsmath,amssymb,amsfonts}
\usepackage{algorithmic}
\usepackage{algorithm}
\usepackage{graphicx}
\usepackage{textcomp}
\usepackage{multicol}
\usepackage{multirow}
\usepackage{xspace}
\usepackage{hyperref}
\newcommand{\mName}{MultiDAG+CL\xspace}
\usepackage{caption}
\usepackage{subcaption}

\usepackage{mathtools}

\title{Curriculum Learning Meets Directed Acyclic Graph \\for Multimodal Emotion Recognition}

\name{Cam-Van Thi Nguyen, Cao-Bach Nguyen, Quang-Thuy Ha, Duc-Trong Le} 

\address{VNU University of Engineering and Technology, Hanoi, Vietnam \\
         \texttt{\{vanntc, 19020218, thuyhq, trongld\}@vnu.edu.vn}\\}
\abstract{
Emotion recognition in conversation (ERC) is a crucial task in natural language processing and affective computing. This paper proposes MultiDAG+CL, a novel approach for Multimodal Emotion Recognition in Conversation (ERC) that employs Directed Acyclic Graph (DAG) to integrate textual, acoustic, and visual features within a unified framework. The model is enhanced by Curriculum Learning (CL) to address challenges related to emotional shifts and data imbalance. Curriculum learning facilitates the learning process by gradually presenting training samples in a meaningful order, thereby improving the model's performance in handling emotional variations and data imbalance. Experimental results on the IEMOCAP and MELD datasets demonstrate that the MultiDAG+CL models outperform baseline models. We release the code for \mName{} and experiments: \url{https://github.com/vanntc711/MultiDAG-CL}.
 \\ \newline  }

\begin{document}

\maketitleabstract

\section{Introduction}\label{sec:intro}

Online social networks' growing popularity has sparked interest in capturing emotions in conversations. Emotion Recognition in Conversation (ERC) has emerged as a critical task in various domains such as chatbots \cite{ghosh-etal-2017-affect}, healthcare \cite{li2019towards}, and social media analysis \cite{polzin2000emotion}. In the field of ERC, researchs can be broadly categorized into unimodal and multimodal approaches. 
Unimodal approaches usually focus on using text as the main modality for emotion recognition. 
Several models have been proposed in the past to tackle unimodal ERC task.
DialogueRNN \cite{majumder2019dialoguernn} introduces a recurrent network to track speaker states and context during the conversation. DialogueGCN \cite{ghosal-etal-2019-dialoguegcn} utilizes graph structures to combine contextual dependencies.

Multimodal Emotion Recognition in Conversation (Multimodal ERC) classifies emotions in conversation turns using text, audio, and visual cues. By incorporating multiple modalities, it provides a comprehensive representation of emotional expressions, including tone of voice, facial expressions, and body language, resulting in improved accuracy and robustness in emotion recognition compared to traditional unimodal ERC approaches.
Several models have been proposed to address the task of multimodal ERC.
The MFN \cite{zadeh2018MFN} synchronizes multimodal sequences using a multi-view gated memory. ICON \cite{hazarika-etal-2018-icon} provides conversational features from modalities through multi-hop memories. The bc-LSTM \cite{poria-etal-2017-context} leverages an utterance-level LSTM to capture multimodal features. MMGCN \cite{hu-etal-2021-mmgcn} uses a graph-based fusion module to capture intra- and inter-modality contextual features. CTNet \cite{lian2021ctnet} utilizes a transformer-based structure to model interactions among multimodal features. CORECT~\cite{nguyen-etal-2023-conversation} leverages relational temporal GNNs with cross-modality interaction support, effectively capturing conversation-level interactions and utterance-level temporal relations.

A Directed Acyclic Graph (DAG) is a directed graph without any directed cycles, comprising vertices and edges, where each edge is directed from one vertex to another, ensuring no closed loops. Building upon this concept, \citet{pmlr-v97-yu19DAGGNN} introduced Directed Acyclic Graph Neural Network (DAG-GNN). Additionally, \citet{shen-etal-2021-directed} presented DAG-ERC, a model combining graph-based and recurrence-based neural architectures to capture information flow in long-distance conversations. However, DAG-ERC's focus has been primarily on unimodal text data, with limited exploration in other modalities.
Curriculum Learning (CL), inspired by human learning, progressively introduces more complex concepts starting from a simple initial state. It establishes a sequence of curricula where the best curriculum with the simplest examples is used to train the classifier in each learning round \cite{bengio2009curriculum, soviany2022curriculum}. 
CL incorporates two key factors: a \textit{difficulty measurer} to assess the difficulty level of training examples, and a \textit{training scheduler} to determine the order of example presentation during training. The difficulty measurer assesses the difficulty level of training examples, while the training scheduler determines the order in which examples are presented to the model during training.
For the ERC task, \citet{yang2022hybrid} proposes a hybrid CL framework specifically for the textual modality only.

\begin{figure*}[ht!]
    \centering
\includegraphics[width=0.86\linewidth]{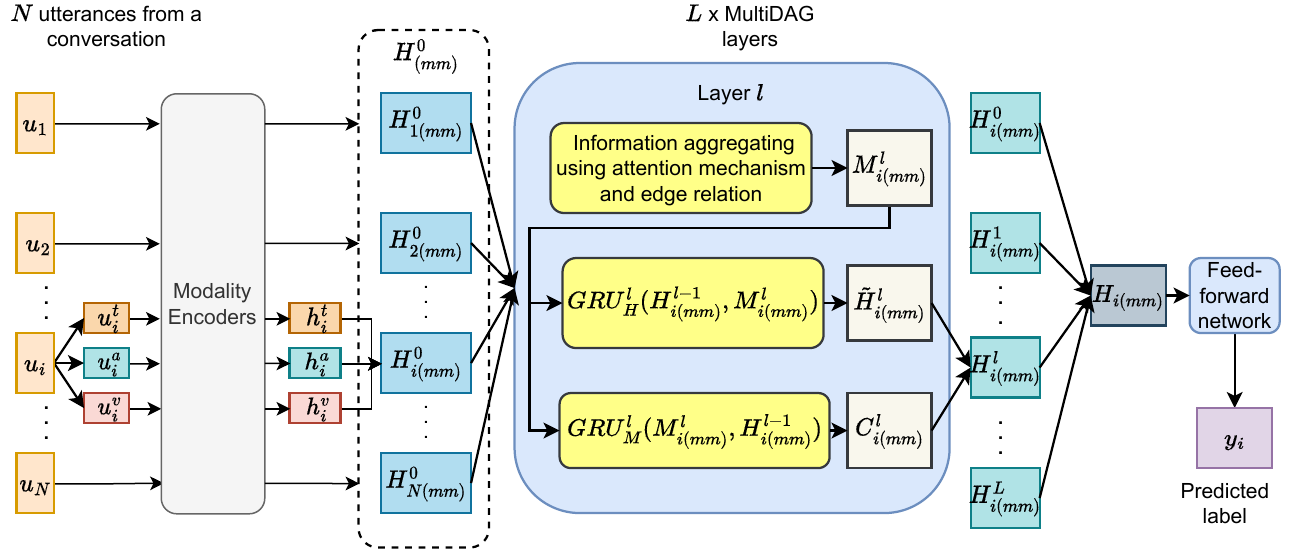}
    \caption{Overall structure of MultiDAG.}
    \label{fig:multiDAG}
\end{figure*}
In this paper, we proposes \textbf{\mName{}}, a multimodal model inspired by DAG-ERC \cite{shen-etal-2021-directed}, designed to overcome the limitations of text-based approaches. It integrates multimodal features using DAG-GNN, enabling a comprehensive understanding of emotions in conversations. Leveraging Curriculum Learning, our model, \textbf{\mName{}}, addresses emotional shift issues and imbalanced data, significantly enhancing ERC model performance on IEMOCAP and MELD datasets. Notably, we are the first to integrate multimodal ERC models with Curriculum Learning strategy. 

\section{Methodology}\label{sec:methods}
Consider a conversation $C$ having utterances $\{u_1, u_2,\dots, u_N\}$ where $N$ is the number of utterances. 
An utterance is a coherent piece of information conveyed by a single participant $p_m$ at a specific moment, where $m \geq 2$. The task of Emotion Recognition in Conversation (ERC) is to predict emotion label of each utterance $u_i$ with predefined emotion label set $E=\{y_1, y_2, \dots, y_r\}$.
Following the multimodal approach, we represent an utterance in terms of three different modalities: audio (a), visual (v), and textual (l). The raw feature representation of utterance $u_i$ is $u_i = \{u_i^{a}, u_i^{v}, u_i^{l}\}$.

\textbf{\mName{}} consists of two core components: \textit{MultiDAG} and \textit{Curriculum Learning-CL}. The \textit{MultiDAG} component represents the model that combines multimodal features without CL integration. The \textit{-CL} component is where Curriculum Learning is incorporated to enhance model performance. 

\subsection{Multimodal ERC with Directed Acyclic Graph - MultiDAG}
\subsubsection{Modality Encoder}
We use modality-specific encoders to generate context-aware utterance feature encoding. For the textual modality, a bidirectional LSTM network captures sequential textual context information, while a Fully Connected Network is used for the acoustic and visual modalities as follows: 
\begin{equation}
    h^a_i = Enc_A(u_i^a) ;  
    h^v_i = Enc_V(u_i^v) ;  
    h^l_i = Enc_L(u_i^l)
\end{equation}
where $Enc_A, Enc_V, Enc_L$ are modality encoder for audio, visual, textual modalities, respectively. These encoders generate the context-aware raw feature encodings $h^a_i, h^v_i, h^l_i$ accordingly. The multimodal feature vector for an utterance $u_{i(mm)}$ corresponding to available modalities is:
\begin{equation}
\label{equation:multimodal}
    H^0_{i(mm)} = h^a_i \oplus h^v_i \oplus h^l_i
\end{equation}

\subsubsection{MultiDAG Construction}
Each utterance in a conversation receives information exclusively from past utterances. This one-way information flow is effectively represented by a Directed Acyclic Graph (DAG), where information moves from predecessors to successors. This characteristic allows the DAG to gather information for a query utterance not only from neighboring utterances but also from more distant ones. Following the multimodal representation input, we initialize the Directed Acyclic Graph Gated Neural Network (DAG-GNN) \cite{pmlr-v97-yu19DAGGNN}. The integration of both remote and local information is executed in a manner analogous to the approach undertaken in DAG-ERC by \citet{shen-etal-2021-directed}. The comprehensive architecture of MultiDAG is visually represented in Figure~\ref{fig:multiDAG}.

At each layer $l$ of the MultiDAG, the hidden state of the utterances is continuously computed from the first utterance to the last utterance. For each utterance $u_{i(mm)}$, the attention weight between $u_{i(mm)}$ and the preceding nodes is calculated by using the hidden state of $u_{i(mm)}$ at layer $l-1$ to attend to the hidden states of the nodes at layer $l$:
\begin{equation}
    a^l_{ij(mm)} = S_{j \in \mathcal{N}_{i(mm)}}(W^l_\alpha[H^l_{j(mm)} \| H^{l-1}_{i(mm)}])
\end{equation}
Here, $S$ denotes the Softmax function; $N_{i(mm)}$ represents the set of preceding nodes leading to $u_i$, $W^l_a$ is a trainable weight matrix, $H^{l-1}_{i(mm)}$ is the hidden state of $u_{i(mm)}$ at layer $l-1$, and $||$ denotes concatenation. 

The attention weight is further utilized in combination with edge relationships to aggregate information.
\begin{equation}
    M^l_{i(mm)} = \sum\limits_{j \in \mathcal{N}_{i(mm)}} a_{ij(mm)}W^l_{r_{ij}}H^l_{j(mm)}
\end{equation}
where $W^l_{r_{ij}} \in \{W^l_0, W^l_1\}$ are trainable parameters. The $0/1$ value represents the edge relationship, distinguishing different or same speakers.

The aggregated information $M^l_{i(mm)}$ interacts with the previous layer's hidden state of $u_{i(mm)}$, $H^{l-1}_{i(mm)}$, through a GRU to generate the final hidden state $\tilde{H}^l_{i(mm)}$ at the current layer:
\begin{equation}
\tilde{H}^l_{i(mm)} = GRU^l_H(H^{l-1}_{i(mm)}, M^l_{i(mm)})
\end{equation}
where $H^{l-1}_{i(mm)}$, $M^l_{i(mm)}$, and $\tilde{H}^l_{i(mm)}$ represent the input, hidden state, and output of the GRU network, respectively. This step is the node information unit. Another GRU serves as the context information unit, modeling the flow of information from the historical context through a layer. In this unit, the roles of $H^{l-1}_i$ and $M^l_i$ in the GRU are exchanged, where $H^{l-1}_{i(mm)}$ controls the propagation of $M^l_{i(mm)}$:
\begin{equation}
C^l_{i(mm)} = GRU^l_M(M^l_{i(mm)}, H^{l-1}_{i(mm)})
\end{equation}

The hidden states of $u_i$ from all layers are concatenated together to create final representation: 
\begin{equation}
    H_{i(mm)} = \|^L_{l=0}(\tilde{H}^l_{i(mm)} + C^l_{i(mm)})
\end{equation}
This representation is then passed through a Feed-Forward Network to perform emotion prediction.
The objective function used to train the model is the cross-entropy loss function. 

\subsection{Curriculum Learning - CL}
 We design a \textbf{Difficulty Measure Function (DMF)} based on the frequency of emotional shift in conversations, and simultaneously construct a \textbf{Training Scheduler} to implement the training process according to the predefined learning curriculum.
 \textbf{\begin{algorithm}[h!]
\algsetup{linenosize=tiny}
\caption{CL Training with DMF}
\label{algorithm:CL_training}
\begin{algorithmic}
\STATE \textbf{Input:} $\mathcal{D}$ - training dataset; $M$ - training model 
\STATE $k$ - number of buckets in baby step scheduler
\STATE $\mathcal{DIF}$ - difficulty measure function
\STATE $t$ - number of epochs; $n$ - number of utterances 
\STATE $e$ - the emotion label of the utterances
\STATE $p(u_i)$ - the speaker’s corresponding utterance $u_i$
\STATE $S$ - Set containing the emotion sequence of speakers; $S[p][i]$ represents the emotion in the $i$-th utterance of speaker $p$
\STATE \textbf{Output:} $M^\ast$ - the optimal model
\STATE $S$ = $\emptyset$, $N_{es} = 0$
\FOR{$i=1$ to $n$} 
    \STATE $S[p[i]] \gets S[p[i]] \cup \{e[i]\}$
\ENDFOR
\STATE $N_{sp}$ = length($S$)
\FOR{$p \in S$}
    \FOR{$i=1$ to length($S[p])-1$}
        \IF{$S[p[i]] \neq S[p[i+1]]$}
            \STATE $N_{shift} \gets N_{shift} + 1$
        \ENDIF
    \ENDFOR
\ENDFOR
\STATE $\mathcal{DIF} = \frac{N_{shift} + N_{sp}}{n + N_{sp}}$
\STATE $\mathcal{D}\prime = sort(\mathcal{D}, \mathcal{DIF})$
\STATE $\mathcal{D}\prime = \{\mathcal{D}^1, \mathcal{D}^2, ..., \mathcal{D}^k\}$ where $\mathcal{DIF}(d_a) < \mathcal{DIF}(d_b), d_a \in \mathcal{D}^i, d_b \in \mathcal{D}^j, \forall{i} < j$
\STATE $\mathcal{D}^{train} = \emptyset$
\FOR{$i=1$ to $t$} 
    \IF {$i \le k$}
    \STATE $\mathcal{D}^{train} = \mathcal{D}^{train} \cup D^i$
    \ENDIF
    \STATE TRAIN($M, \mathcal{D}^{train}$)
\ENDFOR
\STATE return $M\ast$
\end{algorithmic}
\end{algorithm}}

\begin{table*}[ht!]
\centering
\resizebox{2\columnwidth}{!}{%
\begin{tabular}{|c|cccccccc||cc|}
\hline
\multirow{2}{*}{\textbf{Model}} &
  \multicolumn{8}{c||}{\textbf{IEMOCAP}} &
  \multicolumn{2}{c|}{\textbf{MELD}} \\ \cline{2-11} 
 &
  \multicolumn{1}{c|}{Happy} &
  \multicolumn{1}{c|}{Sad} &
  \multicolumn{1}{c|}{Neutral} &
  \multicolumn{1}{c|}{Angry} &
  \multicolumn{1}{c|}{Excited} &
  \multicolumn{1}{c|}{Frustrated} &
  \multicolumn{1}{c|}{Acc. (\%)} &
  w-F1 (\%) &
  \multicolumn{1}{c|}{Acc. (\%)} &
  w-F1 (\%) \\ \hline
bc-LSTM \cite{poria-etal-2017-context} &
  \multicolumn{1}{c|}{33.82} &
  \multicolumn{1}{c|}{78.76} &
  \multicolumn{1}{c|}{56.75} &
  \multicolumn{1}{c|}{64.35} &
  \multicolumn{1}{c|}{60.25} &
  \multicolumn{1}{c|}{60.75} &
  \multicolumn{1}{c|}{60.51} &
  60.42 &
  \multicolumn{1}{c|}{59.62} &
  57.29 \\ \hline
MFN \cite{zadeh2018MFN} &
  \multicolumn{1}{c|}{48.19} &
  \multicolumn{1}{c|}{73.41} &
  \multicolumn{1}{c|}{56.28} &
  \multicolumn{1}{c|}{63.04} &
  \multicolumn{1}{c|}{64.11} &
  \multicolumn{1}{c|}{61.82} &
  \multicolumn{1}{c|}{61.24} &
  61.60 &
  \multicolumn{1}{c|}{60.80} &
  57.80 \\ \hline
ICON \cite{hazarika-etal-2018-icon} &
  \multicolumn{1}{c|}{32.80} &
  \multicolumn{1}{c|}{74.40} &
  \multicolumn{1}{c|}{60.60} &
  \multicolumn{1}{c|}{68.20} &
  \multicolumn{1}{c|}{68.40} &
  \multicolumn{1}{c|}{66.20} &
  \multicolumn{1}{c|}{64.00} &
  63.50 &
  \multicolumn{1}{c|}{58.20} &
  56.30 \\ \hline
DialogueRNN \cite{majumder2019dialoguernn} &
  \multicolumn{1}{c|}{32.20} &
  \multicolumn{1}{c|}{80.26} &
  \multicolumn{1}{c|}{57.89} &
  \multicolumn{1}{c|}{62.82} &
  \multicolumn{1}{c|}{73.87} &
  \multicolumn{1}{c|}{59.76} &
  \multicolumn{1}{c|}{63.52} &
  62.89 &
  \multicolumn{1}{c|}{60.31} &
  57.66 \\ \hline
DialogueGCN \cite{ghosal-etal-2019-dialoguegcn} &
  \multicolumn{1}{c|}{\textbf{51.57}} &
  \multicolumn{1}{c|}{\underline{ 80.48}} &
  \multicolumn{1}{c|}{57.69} &
  \multicolumn{1}{c|}{53.95} &
  \multicolumn{1}{c|}{72.81} &
  \multicolumn{1}{c|}{57.33} &
  \multicolumn{1}{c|}{63.22} &
  62.89 &
  \multicolumn{1}{c|}{58.62} &
  56.36 \\ \hline
DAG-ERC \cite{shen-etal-2021-directed} &
  \multicolumn{1}{c|}{47.59} &
  \multicolumn{1}{c|}{79.83} &
  \multicolumn{1}{c|}{\underline{ 69.36}} &
  \multicolumn{1}{c|}{66.67} &
  \multicolumn{1}{c|}{66.79} &
  \multicolumn{1}{c|}{\textbf{68.66}} &
  \multicolumn{1}{c|}{67.53} &
  68.03 &
  \multicolumn{1}{c|}{61.04} &
  63.66 \\ \hline
MMGCN \cite{hu-etal-2021-mmgcn} &
  \multicolumn{1}{c|}{45.14} &
  \multicolumn{1}{c|}{77.16} &
  \multicolumn{1}{c|}{64.36} &
  \multicolumn{1}{c|}{\underline{ 68.82}} &
  \multicolumn{1}{c|}{\underline{ 74.71}} &
  \multicolumn{1}{c|}{61.40} &
  \multicolumn{1}{c|}{66.36} &
  66.26 &
  \multicolumn{1}{c|}{60.42} &
  58.31 \\ \hline
CTNet \cite{lian2021ctnet} &
  \multicolumn{1}{c|}{51.3} &
  \multicolumn{1}{c|}{79.9} &
  \multicolumn{1}{c|}{65.8} &
  \multicolumn{1}{c|}{67.2} &
  \multicolumn{1}{c|}{\textbf{78.7}} &
  \multicolumn{1}{c|}{58.8} &
  \multicolumn{1}{c|}{68.0} &
  67.5 &
  \multicolumn{1}{c|}{62.0} &
  60.5 \\ \hline
  DAG-ERC+HCL \cite{yang2022hybrid} &
  \multicolumn{1}{c|}{-} &
  \multicolumn{1}{c|}{-} &
  \multicolumn{1}{c|}{-} &
  \multicolumn{1}{c|}{-} &
  \multicolumn{1}{c|}{-} &
  \multicolumn{1}{c|}{-} &
  \multicolumn{1}{c|}{-} &
  \underline{ 68.73} &
  \multicolumn{1}{c|}{-} &
  \underline{ 63.89} \\ \hline 
  COGMEN \cite{joshi-etal-2022-cogmen} &
  \multicolumn{1}{c|}{-} &
  \multicolumn{1}{c|}{-} &
  \multicolumn{1}{c|}{-} &
  \multicolumn{1}{c|}{-} &
  \multicolumn{1}{c|}{-} &
  \multicolumn{1}{c|}{-} &
  \multicolumn{1}{c|}{68.2} &
  67.6 &
  \multicolumn{1}{c|}{-} &
  - \\ \hline\hline
\textbf{MultiDAG (Ours)} &
  \multicolumn{1}{c|}{\underline{ 49.65}} &
  \multicolumn{1}{c|}{79.83} &
  \multicolumn{1}{c|}{66.40} &
  \multicolumn{1}{c|}{67.59} &
  \multicolumn{1}{c|}{71.78} &
  \multicolumn{1}{c|}{\underline{ 67.90}} &
  \multicolumn{1}{c|}{\underline{ 68.30}} & 
  { 68.45} &
  \multicolumn{1}{c|} {\underline{64.29}} & 63.87
   \\ \hline
\textbf{MultiDAG+CL (Ours)} &
  \multicolumn{1}{c|}{45.26} &
  \multicolumn{1}{c|}{\textbf{81.40}} &
  \multicolumn{1}{c|}{\textbf{69.53}} &
  \multicolumn{1}{c|}{\textbf{70.33}} &
  \multicolumn{1}{c|}{71.61} &
  \multicolumn{1}{c|}{66.94} &
  \multicolumn{1}{c|}{\textbf{69.11}} &
  \textbf{69.08} &
  \multicolumn{1}{c|}{\textbf{64.41}} & \textbf{64.00}
   \\ \hline
\end{tabular}%
}
\caption{Performance of approaches on IEMOCAP and MELD datasets. \textbf{Bold} indicates the highest performance, and \underline{underlining} denotes the second-highest. ``-'' represents missing values due to unavailability in original papers.
}
\label{tab:full-results}
\end{table*}
\subsubsection{Difficulty Measure Function (DMF)} 
When designing the difficulty measurement function for a conversation, it is essential to determine what makes a conversation easier or more difficult than others. 
Taking inspiration from \citet{yang2022hybrid}, we constructed a function to calculate the difficulty of a conversation based on the frequency of emotional shift. Here, an emotional shift is defined as occurring when the emotion expressed in two consecutive utterances by the same speaker is different. Specifically, $e(u_i) \ne e(u_k)$, $p(u_i) = p(u_k)$, $\nexists j: i < j < k$, $p(u_i) = p(u_j) = p(u_k)$. Here, $e(u_i)$ and $e(u_k)$ is the emotions of two consecutive utterances
$u_i$ and $u_k$, respectively. 
The more frequent the emotional shift occur in a conversation, the more it is considered difficult. Therefore, the difficulty of $i$-th conversation $c_i$ is as follows:
\begin{equation}
    \mathcal{DIF}(c_i) = \frac{N_{shift}(c_i) + N_{sp}(c_i)}{N_{u}(c_i) + N_{sp}(c_i)}
\end{equation}
where $N_{shift}(c_i)$ and $N_{u}(c_i)$ represent the number of emotional shift in conversation $c_i$ and the total number of utterances in $c_i$, respectively. $N_{sp}(c_i)$ is the number of speakers appearing in conversation $c_i$ and acts as a smoothing factor. 
The algorithm for calculating the difficulty of the conversation is fully described in the Algorithm \ref{algorithm:CL_training}. 

\subsubsection{Training Scheduler} 
The training scheduler is used to organize and schedule the training process by arranging conversations. Specifically, the dataset $\mathcal{D}$ is divided into multiple different bins \{$\mathcal{D}_1, \dots, \mathcal{D}_k$\}, where conversation with similar difficulty are grouped into the same bin. The training process starts with the easiest bin. After training for a certain number of epochs, the next bin is mixed into the current training dataset. Finally, once all bins have been mixed and used, additional epochs of training are performed. 

\section{Experimental Setup}
\subsection{Datasets and Baselines}
We evaluate our approach on the following two ERC datasets: IEMOCAP \cite{busso2008iemocap} and MELD \cite{poria-etal-2019-meld}. 
The detailed statistics of the datasets are reported in Table \ref{tab:data-stat}. 
For the data processing, we use the same split as the work in \cite{hu-etal-2021-mmgcn}. 
We compare our method against several state-of-the-art baselines, including unimodal and multimodal learning approaches.
(Due to the space limit, they are brief described in Section \ref{sec:intro}). The evaluation metrics used are Accuracy (Acc.) and weighted average F1-score (w-F1).
\begin{table}[h!]
\resizebox{\columnwidth}{!}{%
\begin{tabular}{|c|ccc|ccc|c|}
\hline
\multirow{2}{*}{\textbf{Datasets}} & \multicolumn{3}{c|}{\textbf{Conversation}}                     & \multicolumn{3}{c|}{\textbf{Utterances}}                                & \multirow{2}{*}{\begin{tabular}[c]{@{}c@{}}\textbf{Avg.} \\ \textbf{utterances}\end{tabular}} \\ \cline{2-7}
                                   & \multicolumn{1}{c|}{\textbf{Train}} & \multicolumn{1}{c|}{\textbf{Valid}} & \textbf{Test} & \multicolumn{1}{c|}{\textbf{Train}} & \multicolumn{1}{c|}{\textbf{Valid}} & \textbf{Test} &                                                                             \\ \hline
IEMOCAP                            & \multicolumn{2}{c|}{120}                                & 31   & \multicolumn{2}{c|}{5810}                               & 1623 & 66.8                                                                        \\ \hline
MELD                               & \multicolumn{1}{c|}{1038}  & \multicolumn{1}{c|}{114}   & 280  & \multicolumn{1}{c|}{9989}  & \multicolumn{1}{c|}{1109}  & 2610 & 9.57                                                                        \\ \hline
\end{tabular}%
}
\caption{Dataset statistics}
\centering
\label{tab:data-stat}
\end{table}

\subsection{Implementation Details}
We perform hyperparameter tuning for our proposed model on each dataset using hold-out validation with separate validation sets. For the IEMOCAP dataset, the hyperparameter configuration includes a learning rate of 0.0005, a dropout rate of 0.4, 30 epochs of training, and 4 layers of \mName{}. For the MELD dataset, the hyperparameter configuration for the \mName{} model is as follows: a learning rate of 0.00001, a dropout rate of 0.1, 60 epochs of training, and 2 layers of Multi-DAG.

\section{Results and Analysis}
\subsection{Comparision with Baselines}
We conducted a comprehensive comparison of our proposed approach with SOTA multimodal ERC methods, and the results are summarized in Table \ref{tab:full-results}. Due to space constraints, we only report Acc. and w-F1 for the MELD dataset. Our approach, \textit{\mName{}}, which combines the \textit{MultiDAG} model with a curriculum learning strategy, achieves SOTA performance on both the IEMOCAP and MELD datasets. \textit{\mName{}} outperforms previous SOTAs by 1.05\% (DAG-ERC on IEMOCAP) and 0.34\% (DAG-ERC on MELD), respectively. Specifically, our models achieve improvements in individual emotion recognition tasks in most cases, especially for the \textit{Sad}, \textit{Neutral} and \textit{Angry} emotions. In the meantime, we find \textit{Happy}, \textit{Sad}, and \textit{Angry} emotions can be confused with the \textit{Neutral} emotion in some cases (as shown in Fig. \ref{fig:Confusionmatrix}). Such phenomenon is related to imbalanced class distribution.

\subsection{Effect of Modality}
Table \ref{tab:ablation-results} compares the performance of MultiDAG and \mName{} under various multimodal settings on both benchmark datasets. In IEMOCAP, the textual modality performs best among the unimodal settings, while the visual modality shows the lowest results due to noise from factors like camera position and environmental conditions. In bimodal settings, the combination of textual and acoustic modalities performs the best, while the combination of visual and acoustic modalities yields the lowest result. Similar observations are made in the MELD dataset.

\begin{table}[ht!]
\centering
\resizebox{0.8\columnwidth}{!}{%
\begin{tabular}{|c|cl|cl|}
\hline
\multirow{2}{*}{\textbf{Modality}} & \multicolumn{2}{c|}{\textbf{MultiDAG}} & \multicolumn{2}{c|}{\textbf{MultiDAG+CL}} \\ \cline{2-5} 
      & \multicolumn{1}{l|}{IEMOCAP} & MELD & \multicolumn{1}{l|}{IEMOCAP} & MELD \\ \hline
T     & \multicolumn{1}{c|}{68.17}   & 63.66& \multicolumn{1}{c|}{67.12}   &    63.47  \\ \hline
A     & \multicolumn{1}{c|}{49.37}   & 40.27& \multicolumn{1}{c|}{50.58}   &    40.17  \\ \hline
V     & \multicolumn{1}{c|}{33.79}   & 31.27& \multicolumn{1}{c|}{36.69}   &    31.27  \\ \hline \hline
T + A & \multicolumn{1}{c|}{68.42}   & 63.61& \multicolumn{1}{c|}{68.45}   &    63.56  \\ \hline
T + V & \multicolumn{1}{c|}{67.56}   & 63.69& \multicolumn{1}{c|}{67.40}    &    63.62  \\ \hline
A + V & \multicolumn{1}{c|}{52.40}   & 40.54& \multicolumn{1}{c|}{51.86}   &    39.99  \\ \hline \hline
\textbf{T + V + A}                          & \multicolumn{1}{c|}{\textbf{68.45}}   &\textbf{ 63.87}   & \multicolumn{1}{c|}{\textbf{69.08}}     & \textbf{64.00}    \\ \hline
\end{tabular}%
}
\caption{Results of MultiDAG and \mName{} under different modality settings. T, A, V represent the text, audio, visual modality, resepectively.
}
\label{tab:ablation-results}
\end{table}

\subsection{Effect of Curriculum Learning}
The \mName{} model demonstrates notable performance improvement by incorporating curriculum learning for both the IEMOCAP and MELD datasets. 
The effectiveness of curriculum learning relies on factors like the difficulty measure design and training strategy, including the number of buckets in the training set. We perform experiments to select the optimal number of buckets in the CL training scheduler. 
The results shown in the Table \ref{tab:bucket}, indicate that for the IEMOCAP dataset, the optimal number of buckets is 5, while for the MELD dataset, it is 12. 
These findings suggest that the CL strategy is effective in improving the performance of the MultiDAG model on both datasets, with the specific number of buckets tailored to each dataset's representations. 
In summary, our proposed \mName{} model with curriculum learning, significantly contribute to the achieved results.
\begin{table}[!h]
\centering
\resizebox{0.9\columnwidth}{!}{%
\begin{tabular}{|cc|cc|}
\hline
\multicolumn{2}{|c|}{\textbf{IEMOCAP}}          & \multicolumn{2}{c|}{\textbf{MELD}}             \\ \hline
\multicolumn{1}{|c|}{Number of buckets} & w-F1  & \multicolumn{1}{c|}{Number of buckets} & w-F1  \\ \hline
\multicolumn{1}{|c|}{4}                 & 68.05 & \multicolumn{1}{c|}{5}                  & 63.94      \\ \hline
\multicolumn{1}{|c|}{5}                 & \textbf{69.08} & \multicolumn{1}{c|}{8}                & 63.83 \\ \hline
\multicolumn{1}{|c|}{7}                 & 68.84 & \multicolumn{1}{c|}{10}                & 63.89 \\ \hline
\multicolumn{1}{|c|}{10}                & 68.38 & \multicolumn{1}{c|}{12}                & \textbf{64.00} \\ \hline
\multicolumn{1}{|c|}{15}                & 68.36 & \multicolumn{1}{c|}{14}                & 63.96 \\ \hline
\end{tabular}%
}
\caption{Results of \mName{} for different number of buckets in CL training scheduler.}
\label{tab:bucket}
\end{table}

\subsection{Performance for Emotion-shift} 
\begin{figure}[ht!]
\begin{subfigure}{.45\textwidth}
\centering
\includegraphics[width=0.9\linewidth]{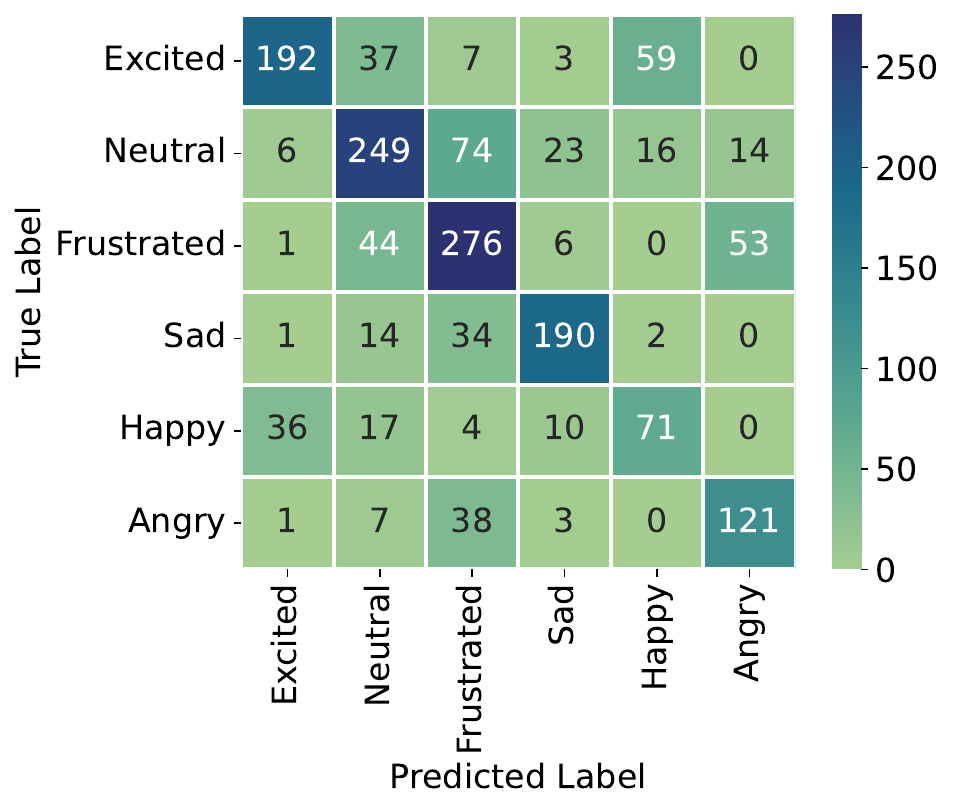}
  \caption{MultiDAG}
  \label{fig:cm1}
\end{subfigure}
\begin{subfigure}{.45\textwidth}
\centering
  \includegraphics[width=0.9\linewidth]{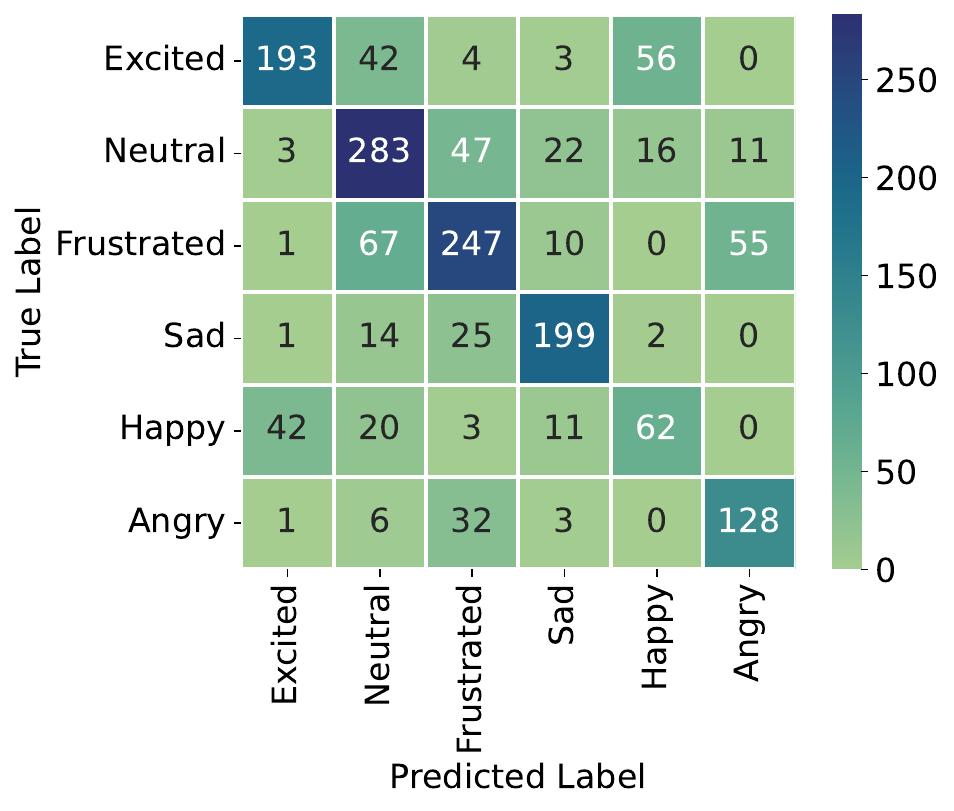}
  \caption{MultiDAG+CL}
  \label{fig:cm2}
\end{subfigure}
\centering
\caption{The confusion matrices on the IEMOCAP.}
\label{fig:Confusionmatrix}
\end{figure}

From the confusion matrices of the MultiDAG and \mName{} models (Figure~\ref{fig:Confusionmatrix}), it can be  observed that the prediction accuracy for the ``Happy'', ``Neutral'', ``Sad'', and ``Angry'' labels is improved when CL is incorporated into the model. Particularly, the misclassification rate of the ``Neutral'' label as ``Disgust'' decreases significantly from 19.3\% in the MultiDAG model to only 12.3\% in \mName{}. 
However, the prediction accuracy for the ``Disgust'' and ``Happy'' labels decreases.
\section{Conclusion}
This paper proposes MultiDAG+CL, a novel approach for Multimodal Emotion Recognition in Conversation that leverages Directed Acyclic Graphs to integrate textual, acoustic, and visual features within a unified framework. The incorporation of Curriculum Learning (CL) addresses challenges related to emotional shifts and data imbalance, enhancing the model's performance.  
Through extensive experiments, we evaluate the performance of both the proposed \mName{} models. Future work includes exploring alternative training schedulers for Curriculum Learning and incorporating a learning-based approach to model emotion label similarity.

\section*{Acknowledgements} 
Cam-Van Thi Nguyen was funded by the Master, PhD Scholarship Programme of Vingroup Innovation Foundation (VINIF), code VINIF.2023.TS147.
\section*{References}
\bibliographystyle{lrec-coling2024-natbib}
\bibliography{lrec-coling2024-example}

\end{document}